\documentclass{article}

\usepackage[preprint]{neurips_2025}
\usepackage{hyperref} 
\usepackage{graphicx}
\usepackage[ruled,vlined]{algorithm2e}
\usepackage{amsmath}
\usepackage{booktabs}
\usepackage{multirow}

\usepackage{amssymb}
\usepackage{CJKutf8}
\usepackage{float}
\usepackage{xcolor}
\usepackage{listings}
\usepackage{color} 
\usepackage{tabularx} 
\usepackage{listings}

\usepackage{graphicx}
\usepackage{adjustbox}

\definecolor{codegreen}{rgb}{0,0.6,0}
\definecolor{codegray}{rgb}{0.5,0.5,0.5}
\definecolor{codepurple}{rgb}{0.58,0,0.82}
\definecolor{backcolour}{rgb}{0.95,0.95,0.92}
\lstdefinestyle{mystyle}{
    backgroundcolor=\color{backcolour},   
    commentstyle=\color{codegreen},
    keywordstyle=\color{magenta},
    numberstyle=\tiny\color{codegray},
    stringstyle=\color{codepurple},
    basicstyle=\footnotesize\ttfamily,
    breakatwhitespace=false,         
    breaklines=true,                 
    captionpos=b,                    
    keepspaces=true,                 
    numbers=left,                    
    numbersep=5pt,                  
    showspaces=false,                
    showstringspaces=false,
    showtabs=false,                  
    tabsize=2,
    language=Python
}
\title{SLOT: Sample-specific Language Model Optimization at Test-time}
\author{Yang Hu$^{1}$\thanks{Equal contribution.} \ \ \ Xingyu Zhang$^{1}$\footnotemark[1] \ \ \ Xueji Fang$^{1}$\ \ \ Zhiyang Chen$^{1}$ \\\textbf{Xiao Wang}$^{2}$ \ \ \ \textbf{Huatian Zhang}$^{3}$ \ \ \ \textbf{Guojun Qi}$^{1}$\thanks{Corresponding author.}\\
$^1$Westlake University \ \ \ $^2$University of Washington \ \ \ $^3$USTC \\
\texttt{\{huyangtorus,xy.zhang042,xiaowang20140001,guojunq\}@gmail.com}\\
\texttt{\{fangxueji,chenzhiyang\}@westlake.edu.cn} \ \ \ \texttt{huatianzhang@mail.ustc.edu.cn}}

\begin{document}

\maketitle

\begin{abstract}

We propose SLOT (Sample-specific Language Model Optimization at Test-time), a novel and parameter-efficient test-time inference approach that enhances a language model's ability to more accurately respond to individual prompts. Existing Large Language Models (LLMs) often struggle with complex instructions, leading to poor performances on those not well represented among general samples.
To address this, SLOT conducts few optimization steps at test-time to update a light-weight sample-specific parameter vector. It is added to the final hidden layer before the output head, and enables efficient adaptation by caching the last layer features during per-sample optimization. By minimizing the cross-entropy loss on the input prompt only, SLOT helps the model better aligned with and follow each given instruction.
In experiments, we demonstrate that our method outperforms the compared models across multiple benchmarks and LLMs. For example, Qwen2.5-7B with SLOT achieves an accuracy gain of 8.6\% on GSM8K from 57.54\% to 66.19\%, while DeepSeek-R1-Distill-Llama-70B with SLOT achieves a SOTA accuracy of 68.69\% on GPQA Diamond among 70B-level models.
Our code is available at \href{https://github.com/maple-research-lab/SLOT}{https://github.com/maple-research-lab/SLOT}.

\end{abstract}

\section{Introduction}

Large Language Models (LLMs) \cite{mann2020language} have shown strong general capabilities in text generation, comprehension, and interaction.
To further boost their performances, test-time scaling has been proposed as a strategy that allocates additional computation during inference to generate more accurate responses for each individual question. In this context, a distinct paradigm known as Test-Time Adaptation (TTA) has emerged, which focuses on optimizing model parameters for individual prompts. Early approaches \cite{sun2020test} implemented instance-specific online adaptation via self-supervision, leveraging loss functions such as entropy loss or reconstruction loss to guide weight updates. More recently, Test-Time Reinforcement Learning (TTRL) \cite{zuo2025ttrl} introduced the use of majority voting results as a reward signal during inference, enabling the model to adapt itself iteratively. However, TTA methods often suffer from high computational overhead due to the need for per-instance updates on large-scale models \cite{akyurek2024surprising}, and designing effective supervision signals for complex LLM tasks remains a significant challenge.

\begin{figure}[H]
    \centering
    \includegraphics[width=1.0\linewidth]{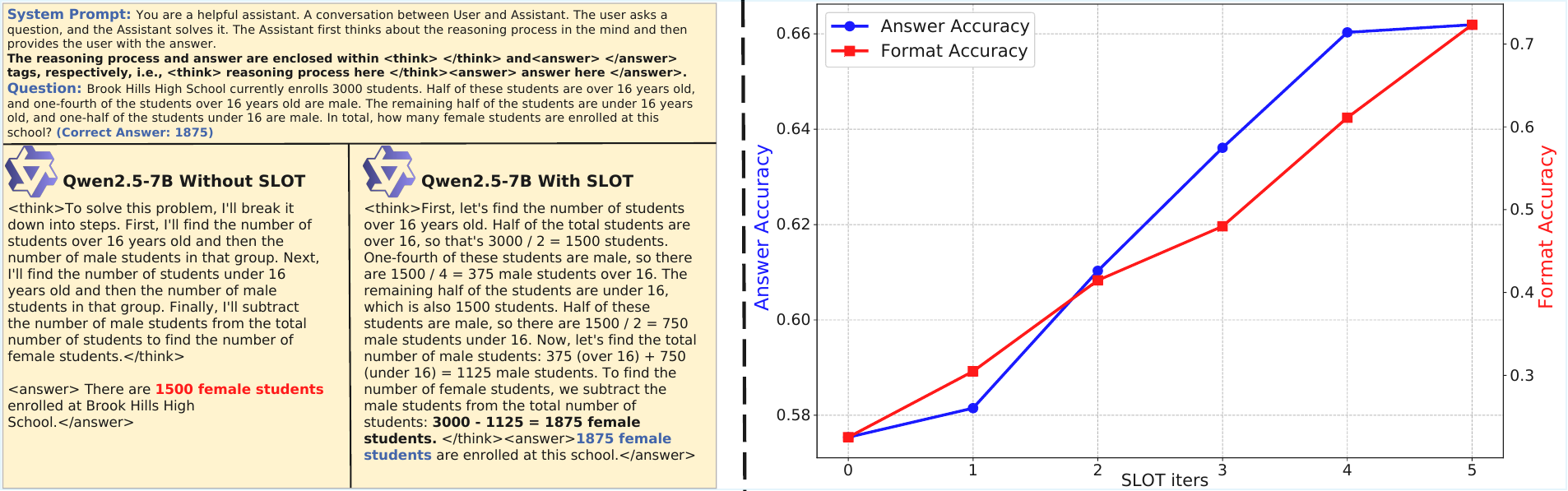}
    \caption{SLOT significantly boosts Qwen-2.5-7B's GSM8K format and answer accuracy at test-time. The \textbf{left} shows the presented prompt and question and compares the output responses from the language model without and with the SLOT used. The \textbf{right} graph shows that both answer accuracy (blue, left axis) and format accuracy (red, right axis) are continuously improved with increasing SLOT optimization iterations from 0 to 5.}
    \label{fig:SLOT_iters}
\end{figure}

The key to effective test-time scaling lies in the method to adapt a pretrained language model to specific inputs. Since such models are trained on general corpora, they cannot be expected to fit all possible prompted instructions, especially those that are not well represented among existing training samples. As shown in Figure~\ref{fig:SLOT_iters}, when applying the popular reasoning template as an additional instruction, Qwen2.5\cite{qwen2025qwen25technicalreport} frequently fails to adhere to the strict formatting requirements and may produce incorrect answers. This behavior likely stems from the model's limited understanding of the presented instruction, due to the absence of such specialized format requirements in its training data. It suggests that the model is underfit in handling unfamiliar, structurally complex instructions.

In this paper, we propose Sample-specific Language Model Optimization at Test-time (SLOT), in order to adapt a pretrained language model to individual prompts at test time. To this end, we regard the prompt itself as a special sample used for supervised training for customized inference. If the model is able to learn and understand the prompt well, the generated answer ought to be better aligned with its instructing content. Particularly, we conduct few optimization steps to minimize the cross-entropy loss on the input prompt only, and use the adapted model to generate the response.

Moreover, to minimize the per-sample overhead and avoid over-adapting model from its original capacity, we use a light-weight additive parameter vector ($\delta$), which only updates the model's final hidden representations right before the output head. This design incurs negligible computing overhead, since it only adapts a single layer of features that can be cached across optimization steps.
We observe that this $\delta$-adapted test-time inference effectively enhances the logits of reasoning-related tokens, encouraging the model to think deeper for more accurate responses.


We conduct experiments on a wide range of LLMs and benchmarks to validate the effectiveness of the SLOT. Particularly, SLOT boosts Qwen2.5-7B's performance on GSM8K by an 8.6\% gain (from 57.54\% to 66.19\%) and enables DeepSeek-R1-Distill-Llama-70B to achieve an accuracy of 68.69\% on GPQA Diamond, a record 3.03\% improvement for 70B opensource models. 
As shown in Figure~\ref{fig:SLOT_iters}, SLOT can successfully improve in both answer accuracy and format accuracy as more optimization steps are adopted.

Our contributions can be summarized as follows.
\begin{itemize} 
    \item  We propose SLOT, a novel test-time training framework to adapt model weights for individual samples. By regarding each input prompt as a special test-time training sample to minimize the per-sample cross-entropy loss, SLOT is able to generate responses more relevant to the input prompt.
    \item SLOT introduces a light-weight sample-specific parameter vector ($\delta$) right before the token prediction head. This design avoids additional forward and backward through the whole language model by caching the features of the last hidden layer, resulting in effective per-sample optimizations.
    \item We conduct extensive experiments on a wide variety of LLMs and benchmarks, and a competitive performance with the SLOT.

\end{itemize}

\section{Related work}

\textbf{In-Context Learning (ICL) and Chain of Thoughts (CoT) } Large language models exhibit ICL capabilities \cite{mann2020language}, where they adapt their behavior based on examples provided within the input prompt, without any gradient updates. The mechanisms underlying ICL are still under investigation \cite{min2022rethinkingroledemonstrationsmakes}. ICL relies solely on the model's forward pass and attention mechanisms operating over the context, while SLOT employs explicit, gradient-based optimization to modify the model's internal state representation for the current input.

\textbf{Parameter-Efficient Fine-Tuning (PEFT)}  PEFT adapts pretrained models for downstream tasks during a \textit{training or fine-tuning phase} by updating only a small number of parameters, keeping the bulk of the LLM frozen. Examples include Adapter Tuning \cite{houlsby2019parameter}, LoRA \cite{hu2022lora}, Prompt Tuning \cite{lester2021power}, Prefix Tuning \cite{li2021prefix}, and P-Tuning v2 \cite{liu2021p}. Crucially, all these methods learn a single set of efficient parameters for an entire task or dataset during a dedicated fine-tuning phase. In contrast, SLOT operates exclusively during the inference phase and optimizes a temporary, sample-specific parameter $\delta$ that is discarded after processing the current input. SLOT does not aim for task-level adaptation but for enhancing the model's processing of the specific instance being evaluated.

\textbf{Test-Time Adaptation (TTA)} TTA aims to address distribution shifts encountered during deployment by dynamically adjusting pre-trained models using only unlabeled test instances, typically in a source-free manner \cite{ xiao2024beyond}. Foundational TTA approaches, often originating in computer vision, include minimizing prediction uncertainty (exemplified by TENT \cite{wang2020tent} which adapts Batch Normalization layers via entropy minimization) and Test-Time Training (TTT) using self-supervision \cite{sun2020test} with tasks like rotation prediction, and later extended to methods like masked autoencoding \cite{gandelsman2022test} and normalizing flows \cite{osowiechi2023tttflow}). While these techniques and associated strategies like combined self-training \cite{su2024towards} or meta-learning for adaptation \cite{sun2023learning} have seen broad application \cite{xiong2023stta, darestani2022test}, their principles are increasingly critical and actively explored for Large Language Models (LLMs). TTA often involves instance-specific online adaptation such as leveraging in-context learning for few-shot reasoning \cite{akyurek2024surprising}, adapting via retrieved data or input rewriting \cite{o2024improving}, or employing Test-Time Reinforcement Learning \cite{zuo2025ttrl}. However, applying TTA to massive LLMs exacerbates general challenges like error accumulation and catastrophic forgetting \cite{liu2021ttt++, zhao2023pitfalls}, and introduces acute issues such as prohibitive computational overhead for per-instance updates \cite{akyurek2024surprising}, the difficulty of designing effective self-supervision for complex linguistic tasks, critical reliance on adaptation data quality and risks to model alignment and fairness. Consequently, developing robust, efficient, lightweight, and safe TTA methods, particularly tailored for LLMs, remains a key ongoing research direction.

\textbf{Test-Time Scaling (TTS)} TTS aims to enhance LLMs' performance, especially on complex reasoning tasks, by allocating additional computational resources during inference, inspired by the benefits of increased `thinking time' processing and often surveyed in works like \cite{sui2025stop, li2025survey, ji2025test}. This typically involves external TTS with fixed pre-trained models \cite{liu2025can}, leveraging inference-time techniques such as sampling-based methods like Self-Consistency (which aggregates multiple Chain-of-Thought paths \cite{wang2022self}) and Best-of-N sampling (often guided by Reward Models \cite{yu2025benchmarking}), or more structured search-based approaches like Tree of Thoughts (ToT) that enable deliberate exploration and backtracking \cite{yao2023tree}. More advanced strategies include adaptive Test-Time Compute, where models dynamically allocate resources based on perceived task difficulty or confidence \cite{manvi2024adaptive, damani2024learning}. A critical insight from TTS research is that strategic inference-time computation can enable smaller LLMs to surpass larger counterparts \cite{liu2025can}, challenging traditional scaling paradigms focused solely on model size. However, TTS methods generally incur significant latency and computational costs, and their efficacy depends heavily on well-calibrated guidance mechanisms, making the balance between performance gains and practical constraints a key area of ongoing investigation.

\section{Approach: Sample-specific Language Model Optimization at Test-time}
\begin{figure}[t]
    \centering
    \includegraphics[width=1.0\linewidth]{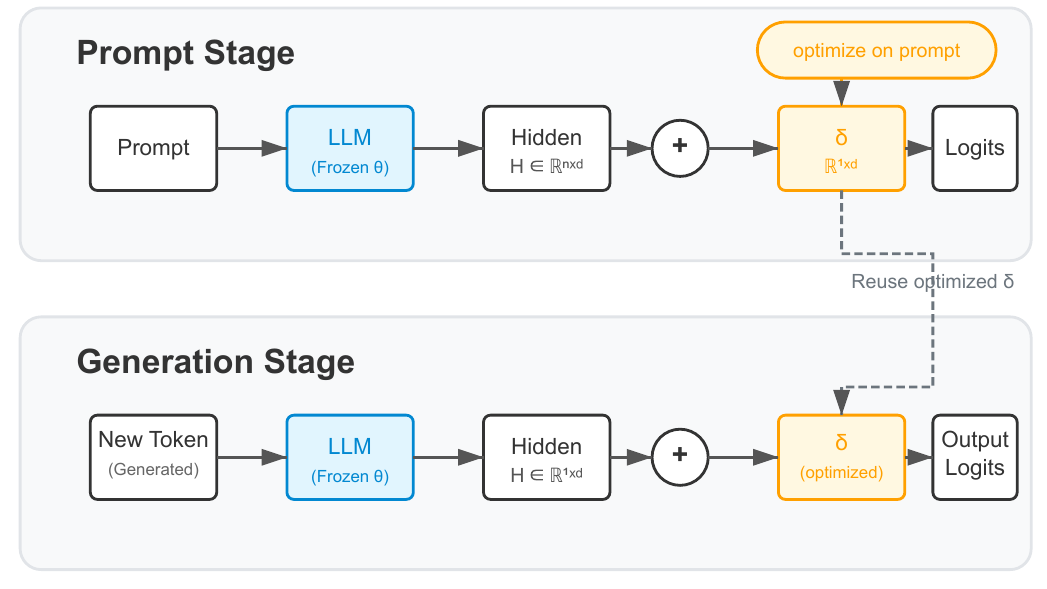}
    \caption{SLOT pipeline during inference. The process consists of two stages: (1) \textbf{Prompt Stage}: Sample-specific parameters $\delta \in \mathbb{R}^{1 \times d}$ are initialized and optimized over $T$ iterations to minimize cross-entropy loss on the input prompt. Each sample in the batch has its own parameters (shared across sequence positions), allowing for efficient adaptation. The original hidden features $H$ are modified by adding $\delta$ to produce $H'$. (2) \textbf{Generation Stage}: During token generation, the optimized $\delta$ parameters are reused without further optimization, modifying the hidden features of newly generated tokens. This approach achieves performance gains with minimal computational overhead, as optimization cost is amortized across the entire generation process. The dashed line illustrates parameter reuse between stages.}
    \label{fig:slot_pipeline}
\end{figure}

\subsection{Problem}
Let $\mathcal{M}$ be a pre-trained language model with parameters $\theta$. 
Formally, given an input sequence of tokens $x = (x_1, x_2, \ldots, x_n)$, the language model processes it to produce the hidden features $H \in \mathbb{R}^{n \times d}$ in an autoregressive manner just before the final linear classifier. Here, $n$ is the sequence length and $d$ is the hidden dimension. 
Then the LM predict the output tokens by computing their probabilities as
\begin{equation}
    p(y|x) = \text{softmax}(W_{\text{LM}}H),
\end{equation}
where $W_{\text{LM}} \in \mathbb{R}^{|V| \times d}$ is the language modeling head weight matrix and $|V|$ is the vocabulary size.

The goal of the proposed SLOT approach is to adapt the trained LM to individual prompts at test-time. To this end, as shown in Fig.~\ref{fig:slot_pipeline}, when a prompt is given, SLOT generates a response with two phases. First, we seek to learn a sample-specific {\em light-weight} parameter $\delta$. Ideally, its size ought to be as small as possible, so that it would not incur heavy computing overhead and can be trained with only a few iterations on individual prompts. We call it \emph{Prompt Stage}. 

Second, we apply $\delta$ to the final hidden features $H$ for the next-token prediction to generate a complete response with the test-time adapted $\delta$. We call it the \emph{Generation Stage}. Below we elaborate on the design details for the model.

\subsection{Method}

To adapt the LLM to a specific sample at test-time for boosting the performance, SLOT introduces a sample-specific parameter $\delta \in \mathbb{R}^{1\times d}$ and performs several test-time optimization steps before generating the response from the input prompt. This parameter is optimized with the input prompt only, and it modifies the final hidden features $H$ into $H'$:
\begin{equation}
H' = H + \delta
\label{eq:apply_delta}
\end{equation}
where $\delta$ is broadcasted across the sequence length dimension ($n$) and added element-wise to the final hidden features $H$. With such a sample-specific weight, the logits computed by the LM head will be changed accordingly: 
\begin{equation}
    \text{logits} = W_{\text{LM}} H' = W_{\text{LM}}(H + \delta).
\end{equation}

\begin{algorithm}[t]
\caption{Sample-specific Language Model Optimization at Test-time (SLOT)}
\label{alg:slot}
\KwIn{Pre-trained language model $\mathcal{M}$ with parameters $\theta$; input token sequence $x = (x_1, \dots, x_n)$; optimization steps $T$; learning rate $\eta$; optimizer hyperparameters $\lambda$.}
\KwOut{Generated sequence extension $y$.}

\tcp{Phase 1: Optimize $\delta$ on the input prompt}
\textbf{Initialize} sample-specific parameter $\delta = \mathbf{0} \in \mathbb{R}^{1 \times d}$\;
Initialize optimizer state (e.g., for AdamW)\;
\For{$t = 0$ \KwTo $T-1$}{
    Compute final hidden features $H = \mathcal{M}_{\text{pre-LM}}(x) \in \mathbb{R}^{n \times d}$ \tcp*{Get features before output head}
    Compute modified hidden features $H' = H + \delta$ \tcp*{Broadcast $\delta$}
    Compute logits $L = W_{\text{LM}}H' \in \mathbb{R}^{n \times |V|}$\;
    Compute loss $\mathcal{L} = \text{CrossEntropyLoss}(L_{:, :-1, :}, x_{2:n})$ \tcp*{Predict next token}
    Compute gradients $g = \nabla_{\delta}\mathcal{L}$\;
    Update $\delta \leftarrow \text{OptimizerStep}(\delta, g, \eta, \lambda)$ \tcp*{e.g., AdamW update}
}
Let $\delta_{\text{opt}} = \delta$ be the final optimized parameter\;

\tcp{Phase 2: Generate using the optimized $\delta$}
Initialize generated sequence $y = ()$ \;
Let current sequence be $x_{\text{current}} = x$\;
\Repeat{end-of-sequence token generated or max length reached}{
    Compute final hidden features for the current token $H_{\text{last}} = \mathcal{M}_{\text{pre-LM}}(x_{\text{current}})[-1, :] \in \mathbb{R}^{1 \times d}$\;
    Compute updated hidden features $H'_{\text{last}} = H_{\text{last}} + \delta_{\text{opt}}$ \tcp*{Reuse optimized $\delta$}
    Compute logits for predicting the next token $L_{\text{next}} = W_{\text{LM}}H'_{\text{last}}$\;
    Sample next token $x_{\text{next}} \sim \text{softmax}(L_{\text{next}})$ \tcp*{Or use greedy decoding}
    Append $x_{\text{next}}$ to $y$\;
    Append $x_{\text{next}}$ to $x_{\text{current}}$\;
}
\Return{y}\;
\end{algorithm}


Although an LM has been fully trained on a dataset consisting of prompts and their answers, we believe its performance on individual prompts can still be boosted further if the model can be adapted at test-time to the scope defined by these input prompts.


Fortunately, a given prompt itself can be regarded as a special training sample for adapting the LM. Particularly, in the \emph{Prompt Stage}, we take several optimization steps to minimize the cross-entropy loss directly on the input prompt, yielding the adjustment $\delta$ that makes the given prompt more `likely' emerging from the adapted LM. Ideally, such a test-time adjustment will adapt the LM to the given prompt so that the generated answers could be more relevantly aligned with the prompt, thereby boosting its success rate in providing correct answers. 

Formally, for each prompt, we first initialize $\delta$ with zeros: $\delta^{(0)} = \mathbf{0} \in \mathbb{R}^{1\times d}$. The zero-initialization ensures that the underlying LM model is not affected at the beginning. Then, we optimize $\delta$ for $T$ steps to minimize the negative log-likelihood (i.e., language modeling loss) on a prompt sequence $x$ of length n,
\begin{equation}
\mathcal{L}(\delta) = -\sum_{i=1}^{n-1} \log p(x_{i+1}|x_{1:i}, \delta)
\label{eq:loss}
\end{equation}
where $p(x_{i+1}|x_{1:i}, \delta)$ is the output probability of the next token $x_{i+1}$ given the context $x_{1:i}$ and the current $\delta$, using the adapted hidden features $H' = H + \delta$. The optimization can be performed with a standard gradient descent optimizer like AdamW \cite{loshchilov2017decoupled}:
\begin{equation}
\delta^{(t+1)} = \text{OptimizerStep}(\delta^{(t)}, \nabla_{\delta} \mathcal{L}(\delta^{(t)}))
\label{eq:update}
\end{equation}
for $t = 0, \dots, T-1$. $\nabla_{\delta} \mathcal{L}(\delta^{(t)})$ denotes the gradient of the loss with respect to $\delta$ at step $t$. 

Note that since $\delta$ is only applied to the final hidden layer in an LM, we can cache the final hidden features $H$. Then in each optimization step, we only need to perform forward and backward with a single linear layer $W_\text{LM}$ upon $H$. This cost is significantly lighter than updating the entire LLM $\theta$, and thus SLOT only incurs a negligible cost with limited memory.

\subsection{Discussions}
Here we would like to discuss some design and potential benefits in SLOT.
\paragraph{Efficiency}
The sample-specific parameter $\delta$ is intentionally applied to the final hidden features before the token prediction head. It provides a short gradient path from the language modeling loss on the input prompt back to $\delta$, resulting in effective few-step optimizations. The parameter $\delta$ is optimized per sample and shared across all token positions in sequentially generating answers. This results in minimal overhead with $d$ parameters, irrespective of sequence length. In the generation phase, it incurs only a negligible $O(d)$ overhead for adding $\delta$ to the final hidden features per token.

\paragraph{Observation}
Adding $\delta$ to the final hidden features before the linear head can be directly interpreted as modulating the output logits for this specific prompt. 
The optimized parameter $\delta$ applies an additive modification $W_{\text{LM}}\delta$ to the logits. To facilitate a detailed analysis of this effect on token probabilities, we formally term this modification the \textbf{Logit Modulation Vector (LMV)}. It is defined as:
\begin{equation}
\text{LMV} \triangleq W_{\text{LM}}\delta \in \mathbb{R}^{|V|}
\label{eq:logit_modulation_vector}
\end{equation}
where $W_{\text{LM}} \in \mathbb{R}^{|V| \times d}$ is the language modeling head matrix, $\delta \in \mathbb{R}^{d}$ is the optimized sample-specific parameter, $|V|$ is the vocabulary size, and $d$ is the hidden dimension. The LMV $\in \mathbb{R}^{|V|}$ thus represents the direct additive shift to the logits for each token in the vocabulary. A positive shift by the LMV corresponds to a strengthened token whose likelihood of being generated is increased, while a negative shift indicates a weakened token with decreased generative likelihood. 

\begin{figure}[t]
    \centering
    \includegraphics[width=0.9\linewidth]{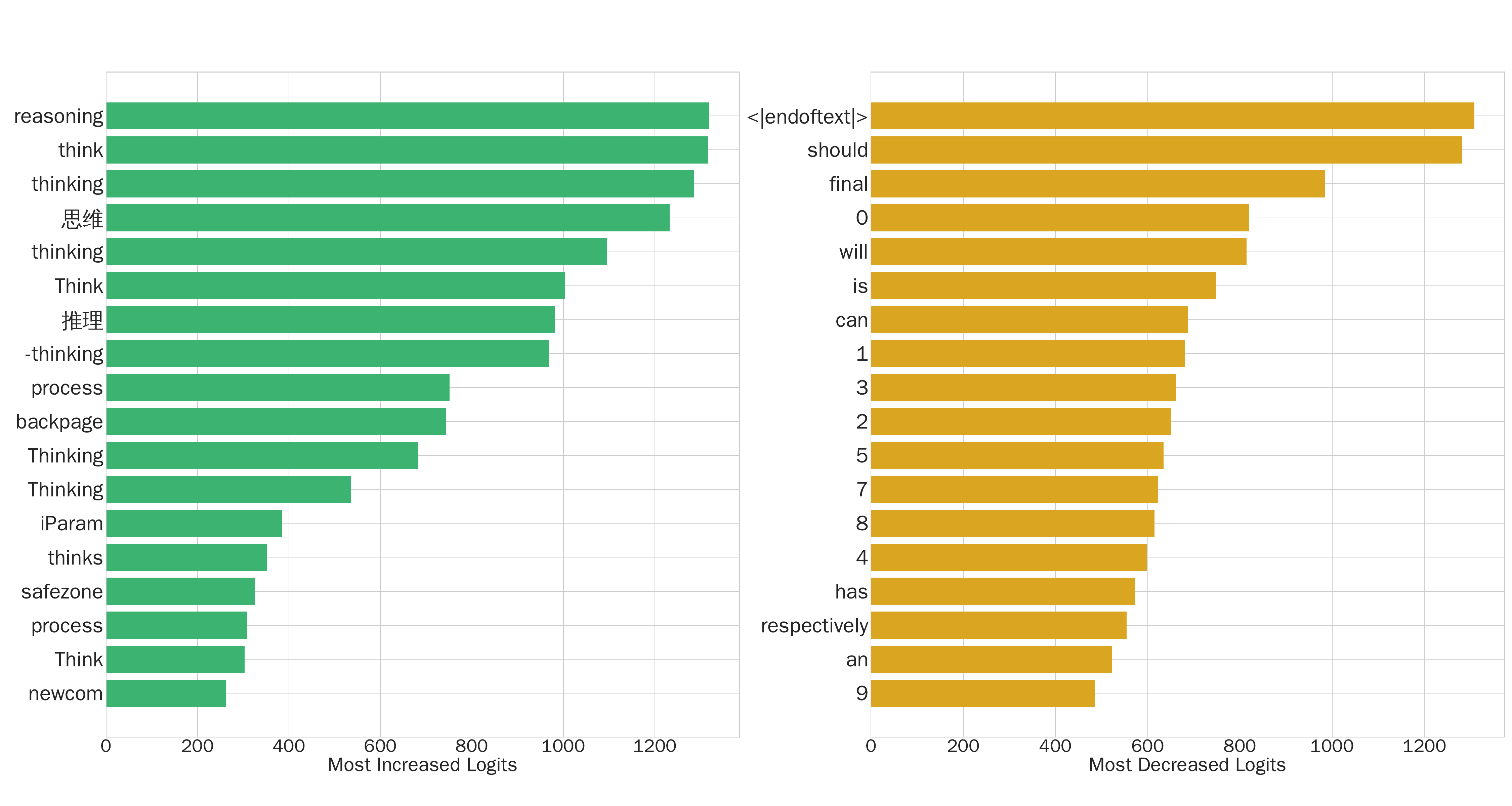}
    \caption{About the most increased and decreased tokens in Qwen-2.5-7B on GSM8K with SLOT(iter=5). We compute the $W_{\text{LM}}(\delta)$ as the degree of change of the logits of the tokens. We find that the most increased tokens are ‘reasoning' related tokens(e.g. reasoning, think, thinking), and the most decreased tokens are numerical tokens(e.g., 0, 1, 2, 3). Also, we find the modal verbs (e.g. should, will, can) is decreased, Figure ~\ref{fig:SLOT_iters}'s left answer example shows, the wrong anser uses many ‘I'll' but don't do actual caculations.}

    \label{fig:increase_decrease}
\end{figure}

We rank tokens with the top increases and decreases in LMV on GSM8K in Figure~\ref{fig:increase_decrease}. It is evident that those tokens related to the reasoning process such as `think' and `reasoning' are significantly enhanced by LMV, which tends to encourage the adapted model to engage in more thorough reasoning. In contrast, numerical tokens, although frequently present in mathematical problems, do not provide sample-specific information and are consequently suppressed, along with common function words such as `should' and `will'. A more interesting finding is the top-1 suppressed token by LMV is the end of text token `$\textless\left|{\rm endoftext} \right|\textgreater$'. This tends to postpone the appearance of the token in the output sequence and thus increase the length of reasoning texts. This may be beneficial in solving challenging problems that require complex reasoning.

\section{Experiment}

In this section, we evaluate the effectiveness of our proposed SLOT method. We compare the performance of baseline Large Language Models (LLMs) against the same models enhanced with SLOT during inference.

\subsection{Implementation Details}
SLOT is implemented with minimal computational overhead relative to standard inference. The optimization of $\delta$ is performed for a small number of iterations (e.g., $T=3$) using the AdamW optimizer. In experiments, we use a learning rate of $\eta=0.01$ and a small weight decay of $1 \times 10^{-8}$, and the epsilon of $1 \times 10^{-5}$  for AdamW.

Zero initialization is adopted with $\delta^{(0)} = \mathbf{0}$. Gradient clipping could be applied, although typically unnecessary for few-step optimization with the chosen learning rate.

\subsection{Models and benchmarks}
We conduct experiments on diverse tasks that involve a variety of LLMs, including
\begin{itemize}
    \item \textbf{Qwen Series:} We consider models in the Qwen family. This includes an earlier model, Qwen-7B \citep{bai2023qwentechnicalreport}, as well as models from the more recent Qwen2.5 generation (Qwen2.5-Math-1.5B, Qwen2.5-Math-7B, Qwen2.5-14B, Qwen2.5-32B) \citep{qwen2025qwen25technicalreport,yang2024qwen2}.
    \item \textbf{Llama Series:} Models in the Llama family are also included for comparison, particularly Llama-3.1-8B and the instruction-tuned Llama-3.1-70B-Instruct \citep{grattafiori2024llama3herdmodels}.
    \item \textbf{DeepSeek Series:} We also evaluate specialized reasoning models from the DeepSeek-R1 series \citep{deepseekai2025deepseekr1incentivizingreasoningcapability}. They are distilled based on Qwen and Llama architectures, spanning various sizes (e.g., DeepSeek-R1-Distill-Qwen-1.5B, 7B, 14B, and 32B; DeepSeek-R1-Distill-Llama-8B, 70B).
\end{itemize}

Model performance is assessed on multiple benchmarks evaluating distinct capabilities:
        \begin{itemize}
            \item \textbf{AIME24}: Derived from the American Invitational Mathematics Examination 2024, testing advanced, competition-style mathematical problem-solving.
            \item \textbf{Math500}: A subset of the MATH dataset \citep{hendrycks2021measuring}, covering diverse mathematical topics from K-12 curriculum to competition levels.
            \item \textbf{GPQA Diamond}: A graduate-level Google-Proof Q\&A benchmark \citep{rein2024gpqa} requiring deep domain knowledge and complex reasoning, primarily in STEM fields.
            \item \textbf{GSM8K}: A dataset of grade school math word problems requiring multi-step reasoning \citep{cobbe2021gsm8k}.
        \end{itemize}
        \begin{itemize}
            \item \textbf{HumanEval}: A standard benchmark for evaluating Python code synthesis from docstrings \citep{chen2021codex}.
        \end{itemize}
        \begin{itemize}
            \item \textbf{C-Eval}: A comprehensive Chinese language evaluation suite covering multiple subjects \citep{huang2023c}. We report performance across its main categories (STEM, Social Science, Humanities, Other), its 'Hard' subset, and the overall average.
        \end{itemize}

We adopt the answer accuracy as the evaluation metric. The baseline performance is obtained from the original models without any test-time adaptation. For the SLOT results, we apply a $T=3$-step optimization with the AdamW solver.

\subsection{Results}

Table \ref{qwen1} presents the comparison between the baseline models and SLOT-enhanced counterparts. The experimental setup follows the protocol used in the official Qwen GitHub repository\footnote{\url{https://github.com/QwenLM/Qwen}} for a fair comparison.

\begin{table}[t!] 
\centering
\caption{Comparison between the baseline and SLOT-enhanced models. SLOT adopts a $T=3$-step adaptation per sample.}
\label{tab:slot_results_scriptsize}
\small
\begin{tabular}{@{}lllrrr@{}} 
\toprule
\textbf{Model} & \textbf{Benchmark} & \textbf{Category} & \textbf{Baseline} & \textbf{With SLOT} & \textbf{Improvement} \\
\midrule
\multirow{8}{*}{Qwen-7B} & \multirow{6}{*}{C-Eval} & STEM & 52.79 & 56.98 & +4.19 \\
                        &                         & Social Science & 79.54 & 79.56 & +0.02 \\
                        &                         & Humanities & 68.60 & 67.39 & -1.21 \\
                        &                         & Other & 59.22 & 57.75 & -1.47 \\
                        &                         & Hard & 36.22 & 44.77 & +8.55 \\ \cmidrule(l){3-6} 
                        &                         & AVERAGE & 62.64 & 63.69 & +1.05 \\
                        \cmidrule(l){2-6} 
                        & GSM8K                   & - & 51.2 & 54.2 & +3.0 \\
                        & HumanEval               & - & 29.9 & 31.7 & +1.8 \\
\bottomrule
\label{qwen1}
\end{tabular}
\end{table}

The results demonstrate the advantage of SLOT over the baseline model at test-time.

For \textbf{Qwen-7B}, SLOT yields notable improvements on several benchmarks. On C-Eval, there is a significant gain of +8.55 points on the `Hard' subset and +4.19 points in the `STEM' category, leading to an overall average improvement of +1.05 points despite minor decreases in `Humanities' and `Other'. Performance gains are also observed on GSM8K (+3.0 points) and HumanEval (+1.8 points), suggesting that the SLOT-enhanced model is beneficial for both reasoning and code generation tasks.

Overall, the results indicate that SLOT is a competitive test-time technique, particularly for improving performance on challenging problems (e.g., C-Eval Hard) and reasoning tasks (e.g., GSM8K). The strong gains in accuracies justify the additional computational overhead incurred by SLOT's per-sample optimization of $\delta$. We will discuss such inference time overhead later, which is quite minor.

\subsection{SLOT for Reasoning}

We also test SLOT on the reasoning benchmarks in open-r1 \footnote{\url{https://github.com/huggingface/open-r1}}). We find that SLOT can boost most results on both the base models and reasoning post-trained models, showing the wide applicability of SLOT in boosting LLMs' performances. 

As reported in Table~\ref{tab:slot_qwen_r1}, SLOT increases the DeepSeek-R1-Distill-Llama-70B baseline on AIME24 from 63.33\% to 73.33\%, and GPQA Diamond from 65.66\% to 68.69\%, achieving the SOTA performance for this task with the open-source  70B baseline model.

\begin{table*}[t!]
  \centering
  \caption{Evaluation with different LLMs on various benchmarks.}
  \label{tab:slot_qwen_r1}
  \resizebox{\textwidth}{!}{%
  \begin{tabular}{@{}llrrrrrrrrr@{}}
    \toprule
    \multicolumn{2}{c}{Model} & \multicolumn{3}{c}{AIME24} & \multicolumn{3}{c}{Math500} & \multicolumn{3}{c}{GPQA Diamond} \\
    \cmidrule(lr){3-5} \cmidrule(lr){6-8} \cmidrule(lr){9-11}
                     &                    & Baseline & wi. SLOT & Improved & Baseline & wi. SLOT & Improved & Baseline & wi. SLOT & Improved \\
    \midrule
    \multirow{6}{*}{Base Model} & Qwen2.5-Math-1.5B    &  6.67 & 10.00 & +3.33 & 43.00 & 43.00 & +0.00 & 27.78 & 27.27 & -0.51 \\
                               & Qwen2.5-Math-7B      & 13.33 & 20.00 & +6.67 & 57.60 & 58.80 & +1.20 & 25.76 & 32.83 & +7.07 \\
                               & Qwen2.5-14B          &  6.67 & 10.00 & +3.33 & 69.80 & 69.60 & -0.20 & 35.86 & 37.88 & +2.02 \\
                               & Qwen2.5-32B          &  3.33 & 13.33 & +10.00 & 65.00 & 66.00 & +1.00 & 36.36 & 42.93 & +6.57 \\
                               & Llama-3.1-8B         &  6.67 & 10.00 & +3.33 & 50.40 & 49.80 & -0.60 & 29.29 & 35.86 & +6.57 \\
                               & Llama-3.1-70B-Instruct & 26.67 & 26.67 & +0.00 & 77.00 & 78.00 & +1.00 & 54.04 & 55.05 & +1.01 \\
    \midrule
    \multirow{6}{*}{Reasoning}   & DeepSeek-R1-Distill-Qwen-1.5B  & 26.67 & 30.00 & +3.33 & 84.40 & 85.40 & +1.00 & 31.82 & 35.35 & +3.53 \\
                               & DeepSeek-R1-Distill-Qwen-7B    & 50.00 & 56.67 & +6.67 & 93.40 & 93.80 & +0.40 & 51.01 & 51.52 & +0.51 \\
                               & DeepSeek-R1-Distill-Qwen-14B   & 66.67 & 73.33 & +6.66 & 95.00 & 95.80 & +0.80 & 60.10 & 61.62 & +1.52 \\
                               & DeepSeek-R1-Distill-Qwen-32B   & 70.00 & 80.00 & +10.00 & 96.80 & 96.20 & -0.60 & 65.66 & 64.65 & -1.01 \\
                               & DeepSeek-R1-Distill-Llama-8B   & 36.67 & 50.00 & +13.33 & 86.80 & 88.00 & +1.20 & 47.47 & 51.52 & +4.05 \\
                               & DeepSeek-R1-Distill-Llama-70B  & 63.33 & 73.33 & +10.00 & 95.80 & 96.00 & +0.20 & 65.66 & 68.69 & +3.03 \\
    \bottomrule
  \end{tabular}
  }
\end{table*}

\subsection{Evaluation on Inference Time}
In Table~\ref{tab:slot_time_overhead_en}, we report the inference time of SLOT with various optimization steps $T$. Note that when SLOT iter $T=0$, it corresponds to the baseline model without any SLOT adaption as $\delta$ is initialized to zero.

As shown in the table, the computational overhead incurred by SLOT with various optimization steps is quite minor compared with the baseline. The time difference between using the baseline and $5$ steps is merely 12.83 seconds, representing only a 7.9\% increase in the inference time.

\begin{table}[t!]
\centering
\caption{Inference time with various SLOT optimization steps. The results are reported based on 30 sampled questions from GSM8K with Qwen-2.5-7B on one NVIDIA-V100 GPU. The average inference time is reported in seconds. The baseline model without SLOT is "iters=0".}
\begin{tabular}{ccccccc}
\toprule

SLOT iters & baseline & 1 & 2 & 3 & 4 & 5 \\
\hline
overall time (s) & 161.49 & 158.72 & 173.93 & 167.07 & 176.03 & 174.32\\
\bottomrule
\end{tabular}
\label{tab:slot_time_overhead_en}
\end{table}

\subsection{Ablation Study}
\label{sec:ablation_hyperparams_ds_1_5b_comprehensive}

We conducted an ablation study on the number of optimization iterations ($T$) and learning rate ($\eta$) for SLOT, using the DeepSeek-R1-Distill-Qwen-1.5B model on the AIME-24 benchmark. As shown in Tab.~\ref{tab:ablation_cross_table_ds_1_5b_comprehensive}, most configurations consistently outperform the original model, indicating that SLOT is relatively insensitive to these hyperparameters. Among the test settings, the configurations of $(T=4, \eta=0.05)$ and $(T=5, \eta=0.05)$ achieve the highest accuracy of 40.00\%, representing a 13.33 percent improvement over the baseline.

In addition to accuracy, we evaluate the inference efficiency of SLOT using two metrics: speed\_input (SI), which measures prompt processing throughput (tokens per second), and speed\_output (SO), which measures generation throughput (new tokens per second) \cite{kwon2023efficient}. Compared to the baseline SI of 12.2 tokens/sec, incorporating a single SLOT optimization iteration results in an SI of 10.62, reflecting a 12\% reduction in prompt processing speed. However, increasing the number of optimization steps does not incur more additional costs since the involved features of the last layer are cached and re-used through the optimization steps. For SO, since SLOT only introduces the addition of a lightweight vector after the final hidden layer, the generation speed is only slightly reduced and remains stable regardless of the number of optimization steps $T$ applied during the prompt stage. 



\begin{table}[!t]
\centering
\scriptsize 
\caption{SLOT hyperparameter ablation on AIME-24 for DeepSeek-R1-Distill-Qwen-1.5B, including accuracy (Acc\%), prompt processing throughput (SI), and generation throughput (SO). For the baseline case ($T=0$), we have an accuracy of 26.67\%, SI of 12.2, SO of 967.84. }
\label{tab:ablation_cross_table_ds_1_5b_comprehensive}
\begin{tabular}{@{}l@{\hspace{2pt}}ccc@{\hspace{4pt}}ccc@{\hspace{4pt}}ccc@{\hspace{4pt}}ccc@{}}
\toprule
\textbf{Iter.} & \multicolumn{3}{c@{\hspace{4pt}}}{\textbf{$\eta=0.01$}} & \multicolumn{3}{c@{\hspace{4pt}}}{\textbf{$\eta=0.05$}} & \multicolumn{3}{c@{\hspace{4pt}}}{\textbf{$\eta=0.1$}} & \multicolumn{3}{c}{\textbf{$\eta=0.2$}} \\
\cmidrule(lr){2-4} \cmidrule(lr){5-7} \cmidrule(lr){8-10} \cmidrule(lr){11-13}
\textbf{($T$)} & Acc\% & SI & SO & Acc\% & SI & SO & Acc\% & SI & SO & Acc\% & SI & SO \\
\midrule
\textbf{1} & 33.33 & 10.62 & 737.08 & 33.33 & 10.13 & 727.70 & 30.00 & 9.64 & 848.40 & 36.67 & 10.15 & 867.84 \\
\textbf{2} & 30.00 & 10.55 & 793.12 & 30.00 & 9.63 & 909.47 & 26.67 & 9.99 & 794.36 & 23.33 & 10.55 & 780.46 \\
\textbf{3} & 26.67 &11.32&804.86 & 30.00 & 13.48 & 967.37 & 36.67 & 10.25 & 786.38 & 33.33 & 10.19 & 789.12 \\
\textbf{4} & 30.00 & 10.49 & 783.06 & \textbf{40.00} & 10.07 & 795.47 & 33.33 & 9.63 & 815.11 & 36.67 & 9.22 & 734.63 \\
\textbf{5} & 23.33 & 9.58 & 856.08 & \textbf{40.00} & 10.10 & 770.47 & 30.00 & 9.90 & 791.13 & 26.67 & 8.43 & 830.09 \\
\bottomrule
\end{tabular}
\end{table}

\section{Conclusion}
We introduced SLOT, a novel test-time adaptation approach that optimizes a lightweight, sample-specific parameter vector added to the final hidden features of large language model. By minimizing the cross-entropy loss on the input prompt itself over a few optimization steps, SLOT efficiently aligns the model more closely with the given instruction, leveraging cached last-layer features to ensure minimal computational overhead. Our experiments demonstrate significant performance improvements on challenging reasoning benchmarks such as GSM8K, AIME24, and GPQA Diamond across various LLMs; for instance, Qwen2.5-7B gained 8.65\% on GSM8K, and DeepSeek-R1-Distill-Llama-70B achieved state-of-the-art GPQA Diamond results for its class. Analysis of the induced logit modulation suggests SLOT encourages deeper reasoning by enhancing the probability of relevant tokens. While SLOT offers a practical method for on-the-fly adaptation, future work might explore adaptive optimization schemes or its efficacy across an even broader range of tasks and modalities.

\bibliographystyle{plain} 
\bibliography{references} 

\newpage

\appendix
\section{Source Code Implementation} 
We present the core part of our source code implementation below in this appendix for our algorithm to facilitate reproducibility of our results.

\begin{lstlisting}[caption={Core implementation snippet for SLOT optimization.}, label={lst:core_code}]
# --- Code related to triggering generation --- 
# Resetting the flag for the *next* sample/question
# Note: The model.generate call likely happens *after* the forward pass shown here completes.
# The logic controlling prompt_only reset should be managed in the inference loop.

os.environ["prompt_only"] = "True" # This likely belongs outside the model's forward pass, before processing a new sample
outputs = model.generate(
     **inputs,
     **generation_params,
)


# --- Code in models' forward ---

## hidden states is the last hidden states
prompt_only = os.environ.get("prompt_only", "False") == "True" 
if prompt_only:
    with torch.enable_grad():
        delta = nn.Parameter(0.0 * torch.randn([1, 1, hidden_states.shape[-1]]).to(hidden_states))
        optimizer = torch.optim.AdamW([delta], lr=0.01, weight_decay=1e-8, eps=1e-5)

        # Optimization loop (T=3 steps)
        for time in range(3):
            optimizer.zero_grad()
            transformed_hidden = hidden_states + delta
            logits = self.lm_head(transformed_hidden)

            # Calculate LM loss on the prompt
            loss_fct = nn.CrossEntropyLoss()
            shift_logits = logits[..., :-1, :].contiguous()
            shift_labels = input_ids[:, 1:].contiguous()
            loss = loss_fct(shift_logits.view(-1, shift_logits.size(-1)), shift_labels.view(-1))

            loss.backward()
            optimizer.step()

        # Store the optimized delta for reuse
        self.delta = delta
        # Apply optimized delta immediately after optimization
        hidden_states = hidden_states + self.delta
        # Set flag to indicate optimization is done for this sample
        os.environ["prompt_only"] = "False"
else:
    # Reuse the stored optimized delta for subsequent generation steps
    hidden_states = hidden_states + self.delta
    
## then the hidden states is processed with lm_head to get logits

\end{lstlisting}

\begin{figure*}[htbp]
    \centering
    \begin{minipage}[b]{0.9\linewidth}
        \centering
        \includegraphics[trim=0 80 0 0, clip,width=1.0\linewidth]{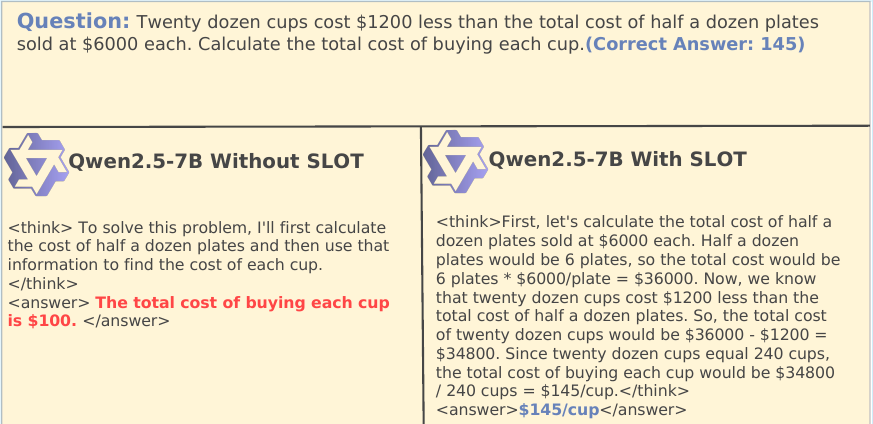}
    \end{minipage}%
    \hfill%
        \centering
    \begin{minipage}[b]{0.9\linewidth}
        \centering
        \vspace{2mm}
        \includegraphics[width=1.0\linewidth]{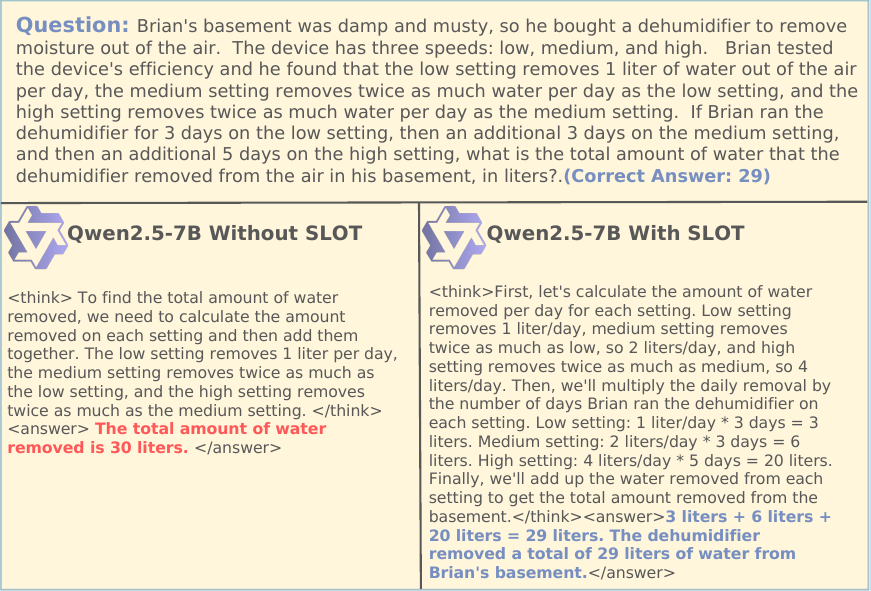}
    \end{minipage}%
    \hfill%
        \centering
    \begin{minipage}[b]{0.9\linewidth}
        \centering
        \vspace{2mm}
        \includegraphics[trim=0 80 0 0, clip,width=1.0\linewidth]{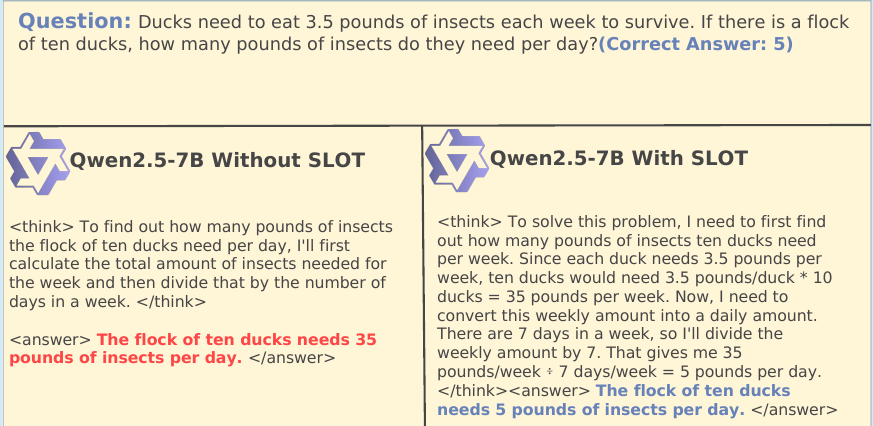}
    \end{minipage}
    \caption{Samples of the comparison between with SLOT and without SLOT in GSM8K. We mark the mistakes in the original responses with red, and the correct answer with blue.}
    \label{fig:all_SLOT_iters}
\end{figure*}

\section{Samples of SLOT}
In Fig.~\ref{fig:all_SLOT_iters}, we demonstrate some examples of model responses. After applying SLOT, Qwen2.5-7B is able to correct the mistakes made in the original responses (marked in red), thus improving model accuracy.

\newpage
\section*{NeurIPS Paper Checklist}

\begin{enumerate}

\item {\bf Claims}
    \item[] Question: Do the main claims made in the abstract and introduction accurately reflect the paper's contributions and scope?
    \item[] Answer: \answerYes{} 
    \item[] Justification: We clearly express our motivation and insight in the abstract and introduction. The contributions are summarized in the introduction.

\item {\bf Limitations}
    \item[] Question: Does the paper discuss the limitations of the work performed by the authors?
    \item[] Answer: \answerNA{} 
    \item[] Justification: The proposed method seems to exhibit no obvious limitations within the research domain discussed in this paper.

\item {\bf Theory assumptions and proofs}
    \item[] Question: For each theoretical result, does the paper provide the full set of assumptions and a complete (and correct) proof?
    \item[] Answer: \answerNA{} 
    \item[] Justification: The paper does not include theoretical results. 

    \item {\bf Experimental result reproducibility}
    \item[] Question: Does the paper fully disclose all the information needed to reproduce the main experimental results of the paper to the extent that it affects the main claims and/or conclusions of the paper (regardless of whether the code and data are provided or not)?
    \item[] Answer: \answerYes{} 
    \item[] Justification: Our paper provides a detailed account of implementation details of our method for reproducing.

\item {\bf Open access to data and code}
    \item[] Question: Does the paper provide open access to the data and code, with sufficient instructions to faithfully reproduce the main experimental results, as described in supplemental material?
    \item[] Answer: \answerNo{} 
    \item[] Justification: We will make all code and pre-trained models available when paper is accepted.

\item {\bf Experimental setting/details}
    \item[] Question: Does the paper specify all the training and test details (e.g., data splits, hyperparameters, how they were chosen, type of optimizer, etc.) necessary to understand the results?
    \item[] Answer: \answerYes{} 
    \item[] Justification: We provide implementation details and experimental setup in Sec. 4.1, Implementation Details and Sec. 4.2, Experimental Setup.

\item {\bf Experiment statistical significance}
    \item[] Question: Does the paper report error bars suitably and correctly defined or other appropriate information about the statistical significance of the experiments?
    \item[] Answer: \answerNo{} 
    \item[] Justification: We evaluate the effectiveness and robustness of the proposed method according to the commonly used practices in the field.

\item {\bf Experiments compute resources}
    \item[] Question: For each experiment, does the paper provide sufficient information on the computer resources (type of compute workers, memory, time of execution) needed to reproduce the experiments?
    \item[] Answer: \answerYes{} 
    \item[] Justification: We specify the type of compute workers in experiments.
    
\item {\bf Code of ethics}
    \item[] Question: Does the research conducted in the paper conform, in every respect, with the NeurIPS Code of Ethics \url{https://neurips.cc/public/EthicsGuidelines}?
    \item[] Answer: \answerYes{} 
    \item[] Justification: We follow the NeurIPS Code of Ethics.

\item {\bf Broader impacts}
    \item[] Question: Does the paper discuss both potential positive societal impacts and negative societal impacts of the work performed?
    \item[] Answer: \answerNA{} 
    \item[] Justification: We believe that our study will not pose any negative societal impacts due to its theoretical nature.
    
\item {\bf Safeguards}
    \item[] Question: Does the paper describe safeguards that have been put in place for responsible release of data or models that have a high risk for misuse (e.g., pretrained language models, image generators, or scraped datasets)?
    \item[] Answer: \answerNA{} 
    \item[] Justification: Our paper does not contain data or models that are at high risk of misuse.

\item {\bf Licenses for existing assets}
    \item[] Question: Are the creators or original owners of assets (e.g., code, data, models), used in the paper, properly credited and are the license and terms of use explicitly mentioned and properly respected?
    \item[] Answer: \answerYes{} 
    \item[] Justification: We cite the datasets used in our paper and give a brief introduction.

\item {\bf New assets}
    \item[] Question: Are new assets introduced in the paper well documented and is the documentation provided alongside the assets?
    \item[] Answer: \answerNA{}  
    \item[] Justification: We do not release new assets.

\item {\bf Crowdsourcing and research with human subjects}
    \item[] Question: For crowdsourcing experiments and research with human subjects, does the paper include the full text of instructions given to participants and screenshots, if applicable, as well as details about compensation (if any)? 
    \item[] Answer: \answerNA{} 
    \item[] Justification: Our paper does not involve human subjects.

\item {\bf Institutional review board (IRB) approvals or equivalent for research with human subjects}
    \item[] Question: Does the paper describe potential risks incurred by study participants, whether such risks were disclosed to the subjects, and whether Institutional Review Board (IRB) approvals (or an equivalent approval/review based on the requirements of your country or institution) were obtained?
    \item[] Answer: \answerNA{} 
    \item[] Justification: Our paper has nothing to do with crowdsourcing and human subjects.

\item {\bf Declaration of LLM usage}
    \item[] Question: Does the paper describe the usage of LLMs if it is an important, original, or non-standard component of the core methods in this research? Note that if the LLM is used only for writing, editing, or formatting purposes and does not impact the core methodology, scientific rigorousness, or originality of the research, declaration is not required.
    \item[] Answer: \answerNA{} 
    \item[] Justification: Our method does not involve LLMs as any important, original, or non-standard components.

\end{enumerate}

\end{document}